\documentclass[11pt]{article}
\usepackage{coling2020}
\usepackage{times}
\usepackage{url}
\usepackage{latexsym}
\usepackage{multirow}
\usepackage{tabularx}
\usepackage{amsmath}
\usepackage{amssymb}
\usepackage{bm}
\usepackage{subfigure}
\usepackage{graphicx}
\usepackage{arydshln}

\colingfinalcopy 

\title{Investigating Catastrophic Forgetting During  \\ Continual Training for Neural Machine Translation}

\author{Shuhao Gu\textsuperscript{\rm 1,2}, Yang Feng\textsuperscript{\rm 1,2}\thanks{\ \ Corresponding author: Yang Feng.}\\ 
\textsuperscript{\rm 1} Key Laboratory of Intelligent Information Processing,\\ Instiute of Computing Technology, Chinese Academy of Sciences (ICT/CAS)\\
\textsuperscript{\rm 2} University of Chinese Academy of Sciences \\
{ \{gushuhao19b,fengyang\}@ict.ac.cn}
}

\date{}

\begin{document}
\maketitle
\begin{abstract}
 
 Neural machine translation (NMT) models usually suffer from catastrophic forgetting during continual training where the models tend to gradually forget previously learned knowledge and swing to fit the newly added data which may have a different distribution, e.g. a different domain. Although many methods have been proposed to solve this problem, we cannot get to know what causes this phenomenon yet. Under the background of domain adaptation, we investigate the cause of catastrophic forgetting from the perspectives of modules and parameters (neurons). The investigation on the modules of the NMT model shows that some modules have tight relation with the general-domain knowledge while some other modules are more essential in the domain adaptation. And the investigation on the parameters shows that some parameters are important for both the general-domain and in-domain translation and the great change of them during continual training brings about the performance decline in general-domain. We conduct experiments across different language pairs and domains to ensure the validity and reliability of our findings.

\end{abstract}

\section{Introduction}

\blfootnote{
    %
    %
    %
    %
    %
    %
     \hspace{-0.65cm}  
     This work is licensed under a Creative Commons 
     Attribution 4.0 International License.
     License details:
     \url{http://creativecommons.org/licenses/by/4.0/}.
}

\begin{figure}[t!]
    \centering
    \includegraphics[width=0.5\columnwidth]{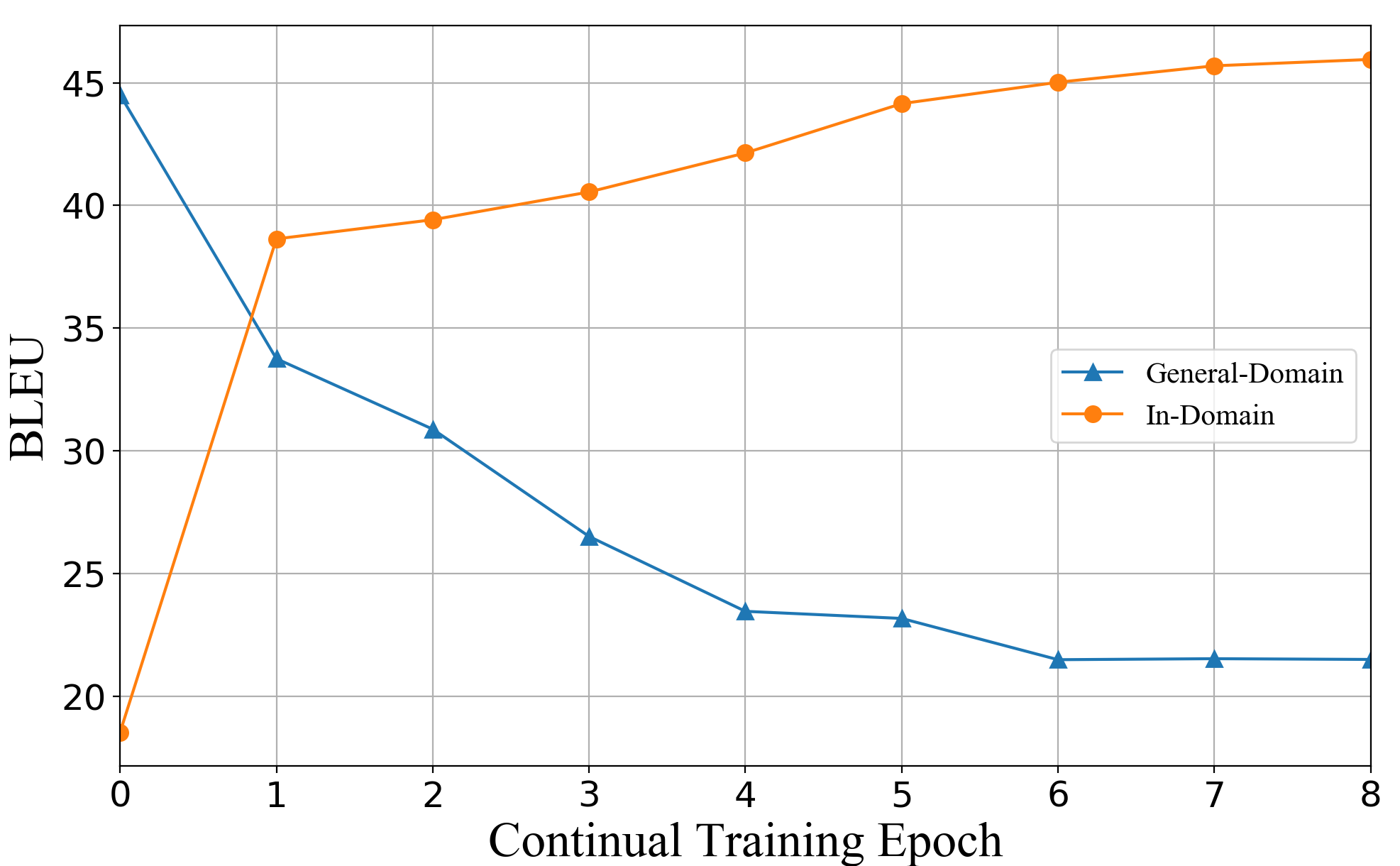}
    \caption{Performance trend for continual training of NMT on the Chinese to English translation task. The model is first trained with the news domain data and then continually trained with the laws domain data (from epoch 0). The y-axis represents the BLEU scores on the test sets of the corresponding domains.}
    \label{fig:CF}
\end{figure}

Neural machine translation (NMT) models~\cite{kalchbrenner2013recurrent,ChoMGBBSB14,sutskever2014sequence,BahdanauCB14,GehringAGYD17,VaswaniSPUJGKP17} have achieved state-of-the-art results and have been widely used in many fields. Due to numerous parameters, NMT models can only play to their advantages based on large-scale training data. However, in practical applications, NMT models often need to perform translation for some specific domain with only a small quantity of in-domain data available. 
In this situation, continual training~\cite{luong2015stanford}, which is also referred to as fine-tuning, is often employed to improve the in-domain translation performance. In this method, the model is first trained with large-scale general-domain training data and then continually trained with the in-domain data. With this method, the in-domain performance can be improved greatly, but unfortunately, the general-domain performance decline significantly, since NMT models tend to overfit to frequent observations (e.g. words, word co-occurrences, translation patterns) in the in-domain data but forget previously learned knowledge. This phenomenon is called catastrophic forgetting. Figure~\ref{fig:CF} shows the performance trends on the in-domain and general-domain.

Many methods have been proposed to address the catastrophic forgetting problem under the scheme of fine-tuning. \newcite{FreitagA16} ensembles the general-domain model and the fine-tuned model together so that the integrated model can consider both domains. 
\newcite{dakwale2017fine} introduces domain-specific output layers for both of the domains and thus the domain-specific features of the two domains can be well preserved. 
\newcite{ThompsonGKDK19}, \newcite{BaroneHGS17}, and \newcite{KhayrallahTDK18} propose regularization-based methods that introduce an additional loss to the original objective to help the model trade off between the general-domain and in-domain. All these methods show their effectiveness and have mitigated the performance decline on general-domain, but we still don't know what happened inside the model during continual training and why these methods can alleviate the catastrophic forgetting problem. The study on these can help to understand the working mechanism of continual training and inspire more effective solutions to the problem in return.

Given above, in this paper, we focus on the catastrophic forgetting phenomenon and investigate the roles of different model parts during continual training. To this end, we explore the model from the granularities of modules and parameters (neurons). In the module analyzing experiments, we operate the model in two different ways, by freezing one particular module or freezing the whole model except for this module. We find that different modules preserve knowledge for different domains. In the parameter analyzing experiments, we erase parameters according to their importance which is evaluated by the Taylor expansion-based method~\cite{MolchanovTKAK17} . According to the experimental results, we find that some parameters are important for both of the general-domain and in-domain and meanwhile they change greatly during domain adaptation which may result in catastrophic forgetting.  To ensure the validity and reliability of the findings, we conduct experiments over different language pairs and domains. 

Our main contributions are summarized as follows:
\begin{itemize}
\setlength{\itemsep}{0pt}
\setlength{\parsep}{0pt}
\setlength{\parskip}{0pt}
\item We propose two analyzing methods to explore the model from the perspectives of modules and parameters, which can help us understand the cause of catastrophic forgetting during continual training. 
\item We find that some modules tend to maintain the general-domain knowledge while some modules are more essential for adapting to the in-domain.
\item We find that some parameters are important for both of the general-domain and in-domain, and their over-change in values may result in performance slipping. 
\end{itemize}

\section{Background}

In our work, we apply our method under the framework of \textit{Transformer}~\cite{VaswaniSPUJGKP17} which will be briefly introduced here. We will also introduce the terminology and symbols used in the rest of the paper.
We denote the input sequence of symbols as $\mathbf{x}=(x_1,\ldots,x_J)$, the ground-truth sequence as $\mathbf{y}^{*}=(y_1^{*},\ldots,y_{I^*}^{*})$ and the translation as $\mathbf{y}=(y_1,\ldots,y_I)$.

 \textbf{The Encoder} 
The encoder is composed of $\mathnormal{N}$ identical layers. Each layer has two sublayers. The first is a multi-head {\em self-attention sublayer} (abbreviated as SA) and the second is a fully connected feed-forward network, named {\em FFN sublayer} (abbreviated as FFN). Both of the sublayers are followed by a residual connection operation and a layer normalization operation. The input sequence $\mathbf{x}$ will be first fed into the {\em embedding layer} (abbreviated as Emb) and converted to a sequence of vectors $\mathbf{E}_x=[E_x[x_1];\ldots;E_x[x_J]]$ where $E_x[x_j]$ is the sum of word embedding and position embedding of the source word $x_j$. 
Then, this input sequence of vectors will be fed into the encoder and the output of the $\mathnormal{N}$-th layer is taken as source hidden states and we denote it as $\mathbf{H}$. 

\textbf{The Decoder}
The decoder is also composed of $\mathnormal{N}$ identical layers. In addition to the same kinds of two sublayers in each encoder layer, a third sublayer is inserted between them, named {\em cross-attention sublayer} (abbreviated as CA), which performs multi-head attention over the output of the encoder stack. The final output of the $\mathnormal{N}$-th layer gives the target hidden states, denoted as $\mathbf{S}=[\mathbf{s}_1;\ldots;\mathbf{s}_I]$, where $\mathbf{s}_i$ is the hidden states of $y_i$. 

\textbf{The Objective} 
The output of the decoder will be fed into the {\em output layer} (abbreviated as Out) and we can get the predicted probability of the $i$-th target word over the target vocabulary by performing a linear transformation and a softmax operation to the target hidden states:
\begin{equation}
    p(y_i | \mathbf{y}_{<i}, \mathbf{x}) \propto \exp({\mathbf W}_o  {\mathbf s}_i + \mathbf{b}_o)
\end{equation}
where ${\mathbf W}_o \in \mathbb{R}^{d_{model}\times|\mathrm{V}_t|}$ is the parameter matrix of the {\em output layer}, and $|\mathrm{V}_t|$ is the size of target vocabulary. 
The model is optimized by minimizing a cross-entropy loss which maximizes the probability of the ground-truth sequence with the teacher-forcing training:
\begin{equation}\label{eq::loss}
    \mathcal{L} = -\frac{1}{I}\sum_{i=1}^I \log p(y_i^{*} | \mathbf{y}_{<i}, \mathbf{x})
\end{equation}
where $I$ is the length of the target sentence. A detailed description can be found in~\newcite{VaswaniSPUJGKP17}.

\section{Module Analysis}
In this work, we will investigate the cause of the catastrophic forgetting phenomenon from the perspectives of modules and parameters. As we know, the structure and the various kinds of modules in it has a large impact on translation. In this section, therefore, we will study the function of different modules by isolating them during continual training. The study on the parameters will be discussed in Section~\ref{par}.


\subsection{Analyzing Strategies}
We propose two training strategies during the continual training process. The first is to freeze the target module but update the rest of the model, called module-frozen training; the second, in contrast, is to update the target module but freeze the rest of the model, called module-updated training. In this way, we can estimate the ability of each module for preserving the general-domain knowledge and for adapting to the in-domain.
In addition, we group the target module based on two criteria. The first is based on its position, e.g., we freeze or update all the sublayers in the {\em first two layers} of the encoder or the {\em last two layers} of the decoder; the second is based on its type, e.g., we freeze or update the {\em self-attention sublayers} or the {\em cross-attention sublayers} in all the decoder layers.


\subsection{Experiments}\label{exp}
\subsubsection{Data Preparing}
We conduct experiments on the following data sets across different languages and domains. 

\textbf{Chinese$\rightarrow$English}. For this task, general-domain data is from the LDC corpus\footnote{The corpora include LDC2002E18, LDC2003E07, LDC2003E14, Hansards portion of LDC2004T07, LDC2004T08, and LDC2005T06.} that contains 1.25M sentence pairs. The LDC data is mainly related to the \textbf {News} domain. MT06 and MT02 are chosen as the development and test data, respectively. We choose the parallel sentences with the domain label \textbf {Laws} from the UM-Corpus~\cite{TianWCQOY14} as our in-domain data. We filter out repeated sentences and chose 206K, 2K, and 2K sentences randomly as our training, development, and test data, respectively. We tokenize and lowercase the English sentences with Moses\footnote{http://www.statmt.org/moses/} scripts. For the Chinese data, we perform word segmentation by using Stanford Segmenter\footnote{https://nlp.stanford.edu/}.

\textbf{English$\rightarrow$French}. For this task, we choose 600K sentences randomly from the WMT 2014 corpus as our general-domain data, which are mainly related to the \textbf{News} domain. We choose newsdev2013 and newstest2013 as our development and test data, respectively. The in-domain data with 53K sentences is from WMT 2019, and it is mainly related to the \textbf{Biomedical} domain. We choose 1K and 1K sentences randomly from the corpora as our development and test data, respectively. We tokenize and truecase the corpora.

\textbf{English$\rightarrow$German}. For this task, the general-domain data is from the WMT 2016 English to German translation task which is mainly \textbf {News} texts. It contains about 4.5M sentence pairs. We choose the news-test 2013 for validation and news-test 2014 for the test. For the in-domain data, we use the parallel training data from the IWSLT 2015 which is mainly from the \textbf {Spoken} domain. It contains about 194K sentences. We choose the 2012dev for validation and 2013tst for the test. We tokenize and truecase the corpora.

Besides, integrating operations of 32K, 16K, and 30K are performed to learn BPE~\cite{SennrichHB16a} on the general-domain data and then applied to both the general-domain and in-domain data. The dictionaries are also built based on the general-domain data.

\subsubsection{Systems\label{setting}}
We use the open-source toolkit called {\em Fairseq-py}~\cite{ott2019fairseq} released by Facebook as our Transformer system.
We train the model with two sets of parameters. For the quantitatively analyzing experiments, the system is implemented as the base model configuration in~\newcite{VaswaniSPUJGKP17} strictly. For the visualizing experiments, we employ a {\em tiny} setting: the embedding size is set to 32, the FFN size is set to 64, and the rest is the same with the base model.

\begin{figure}[t]
    \centering
    \includegraphics[width=\columnwidth]{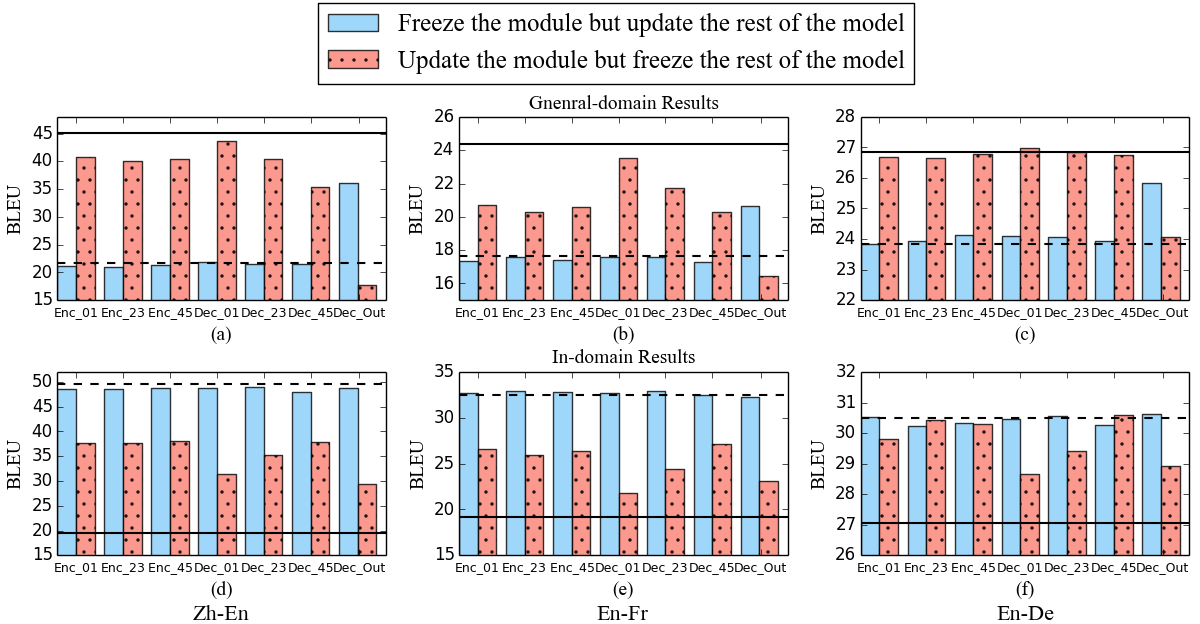}
    \caption{The BLEU of the experiments based on the position of the target module when only freezing the target module (left blue bars) or only updating the target module (right red bars). The first three subfigures (a-c) represent the general-domain BLEU and the last three subfigures (d-f) represent the in-domain BLEU. The horizontal solid line in each subfigure denotes the BLEU before continual training while the dashed line denotes the BLEU after continual training. The x-axis denotes the module at different positions, e.g., 'Enc\_01' denotes the first two layers of the encoder, 'Dec\_out' denotes the output layer of the decoder. }
    \label{fig:pos}
\end{figure}

\subsubsection{The Results Based on the Position of the Target Module}
The results of the experiments, which freeze or update the target module based on its position, are shown in Figure~\ref{fig:pos}, where the left blue bars correspond to the results of the module-frozen experiments and the right red bars correspond to the results of module-updated experiments. 
As can be seen from the results of the module-frozen experiments, for different positions of the encoder and decoder, freezing any single module has a small impact on both the general-domain and in-domain BLEU, when compared with the normal continual training. However, freezing the output layer of the decoder can better alleviate the catastrophic forgetting phenomenon without degrading the in-domain BLEU, which implies that the output layer is more capable of maintaining general-domain knowledge. 
As for the results of the module-updated experiments, firstly, only updating the encoder can bring larger improvements on the in-domain compared with only updating the decoder, which implies that the encoder is more essential for adapting to the in-domain. Secondly, higher layers of the decoder tend to adapt to the in-domain, which, however, is not the case in the encoder. Lastly, only updating the output layer results in a bad performance on both domains, which indicates that the output layer highly depends on its surrounding modules.

\begin{figure}[t]
    \centering
    \includegraphics[width=\columnwidth]{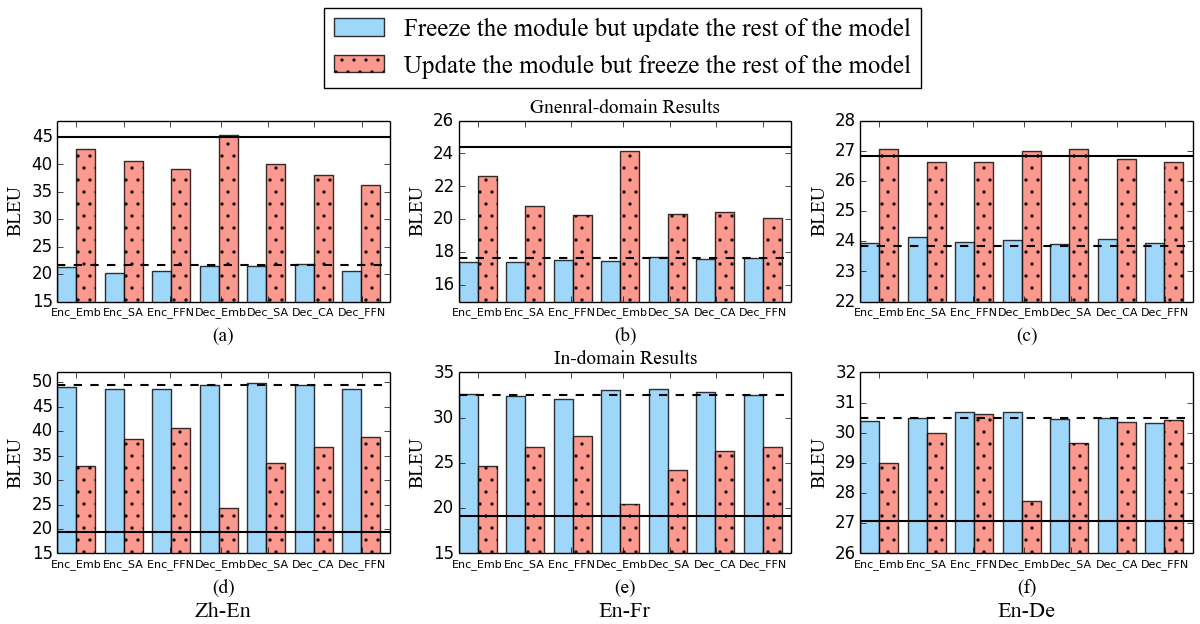}
    \caption{The BLEU of the experiments based on the type of the target module. The x-axis denotes the module of different types, where 'Emb', 'SA', 'CA', and 'FFN' denote the embedding layer, self-attention sublayer, cross-attention sublayer, and FFN sublayer, respectively. The rest of the denotations is just the same as in Figure~\ref{fig:pos}.}
    \label{fig:type}
\end{figure}

\subsubsection{The Results Based on the Type of the Target Module}
Figure~\ref{fig:type} shows the results of the experiments based on the type. 
The results of the module-frozen experiments (the left solid bars) show that freezing any type of module has little effect on the final results. As for the results of the module-updated experiments (the right dotted bars), we find that both the encoder embedding layer and the decoder embedding layer tend to preserve the general-domain knowledge, meanwhile, they are bad at adapting to the in-domain. In contrast, the FFN layers are more essential for adapting to the in-domain, though it is also easier for them to cause the catastrophic forgetting problem in general-domain. 

\subsection{Summary and Inspiration}
In this section, we analyze the impacts of different modules on the general-domain and in-domain translation during continual training. we find that some modules, e.g., the output layer, the embedding layers, tend to preserve the general-domain knowledge; some modules, e.g., the encoder layers, the FFN layers, are more essential for adapting to the in-domain. Inspired by our findings, we can freeze those modules which are more important for general-domain during continual training to avoid catastrophic forgetting. To reduce the potential loss on the in-domain translation, we can extend the size of those frozen layers or add domain-specific layers parallel with them, and only update those newly added parameters during continual training.

\section{Parameter Analysis}\label{par}


When the structure of the NMT model is fixed, the performance is determined by its parameters. During continual training, the change of the training data distribution makes the distribution of model parameters vary accordingly, which leads to the variation of translation performance on both the general-domain and in-domain. 
Motivated by this, therefore, we want to investigate the changing trend of model parameters during continual training, aiming to figure out the influence of different parameters. Intuitively, different parameters may have different importance for the NMT model, thus we firstly propose a method for evaluating the importance of parameters in this section. Then, we erase the model parameters increasingly according to the importance to see the change of BLEU scores. Finally, we visualize the parameters and measure their variations in the values to establish the connection between the parameter variation and the catastrophic forgetting phenomenon.

\begin{figure}[t]
    \centering
    \includegraphics[width=\columnwidth]{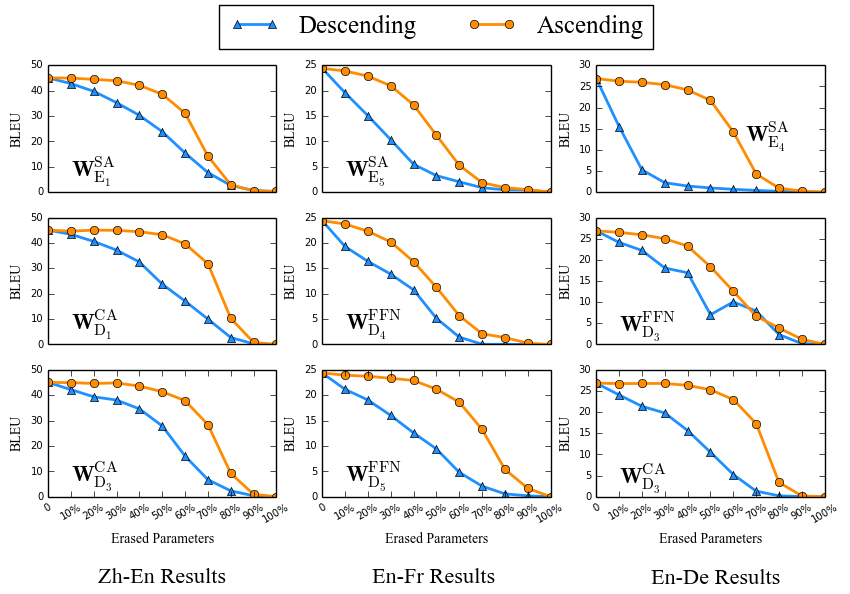}
    \caption{BLEU scores of the parameter erasure experiments with the model trained and tested on the general-domain. The blue curve with triangles stands for erasing parameters in descending order according to the importance while the orange curve with dots stands for erasing parameters in ascending order. }
    \label{fig:erasure1}
\end{figure}

\subsection{Importance Evaluation Method}\label{sec:method}
To evaluate the importance of each parameter, we adopt a criterion based on the Taylor expansion~\cite{MolchanovTKAK17}, where we directly approximate the change in loss when removing a particular parameter. Let $h_i$ be the output produced from parameter $i$ and $H$ represents the set of other parameters. 
Assuming the independence of each parameter in the model, the change of loss when removing a certain parameter can be represented as:
\begin{equation}
    |\Delta\mathcal{L}(h_i)| = |\mathcal{L}(H, h_i=0) - \mathcal{L}(H, h_i)|
\end{equation} 
where $\mathcal{L}(H, h_i=0)$ is the loss value if the parameter $i$ is pruned and $\mathcal{L}(H, h_i)$ is the loss if it is not pruned. For the function $\mathcal{L}(H, h_i)$, its Taylor expansion at point $h_i = a$ is:
\begin{equation}
  \mathcal{L}(H, h_i) = \sum_{n=0}^{N}\frac{\mathcal{L}^{n}(H, a)}{n!}(h_i - a)^n + R_N(h_i)
\end{equation}
where $\mathcal{L}^{n}(H, a)$ is the $n$-th derivative of $\mathcal{L}(H, h_i)$ evaluated at point $a$ and $R_n(h_i)$ is $n$-th remainder. 
Then, approximating $\mathcal{L}(H, h_i=0)$ with a first-order Taylor polynomial where $h_i$ equals zero, we get:
\begin{equation}
    \mathcal{L}(H, h_i=0) = \mathcal{L}(H, h_i)-\frac{\partial \mathcal{L}(H, h_i)}{\partial h_i}h_i-R_1(h_i)
\end{equation}
The remainder $R_1$ can be represented in the form of Lagrange:
\begin{equation}
    R_1(h_i) = \frac{\partial^2\mathcal{L}(H, h_i)}{\partial^2\delta h_i}h_i^2,
\end{equation}
where $\delta \in (0,1)$. Considering the use of ReLU activation function in the model, the first derivative of loss function tends to be constant, so the second order term tends to be zero in the end of training. Thus, we can ignore the remainder and get the importance evaluation function as follows:
\begin{equation}
    \Theta_\mathrm{TE}(h_i) = \left|\Delta\mathcal{L}(h_i)\right|  =  \left|\frac{\partial \mathcal{L}(H, h_i)}{\partial h_i}h_i\right|
\end{equation} 
Intuitively, this criterion disvalues the parameters that have an almost flat gradient of the objective function. In practice, we need to accumulate the product of the activation and the gradient of the objective function w.r.t to the activation, which is easily computed during back-propagation. Finally, the evaluation function is shown as:
\begin{equation}\label{eq:te}
    \Theta_\mathrm{TE}(h_i^l)=\frac{1}{T}\sum_{t}\left|\frac{\delta\mathcal{L}(H, h_i^l)}{\delta h_i^l}h_i^l\right|,
\end{equation}
where $h_i^l$ is the activation value of the $i$-th parameter of $l$-th module and $T$ is the number of the training examples. The criterion is computed on the all training data and averaged over $T$. 

  \begin{figure}[t]
    \centering
    \subfigure[Zh-En $\mathbf{W}_{\mathrm{E}_0}^{\mathrm{SA}}$]{
        \includegraphics[width=0.3\columnwidth]{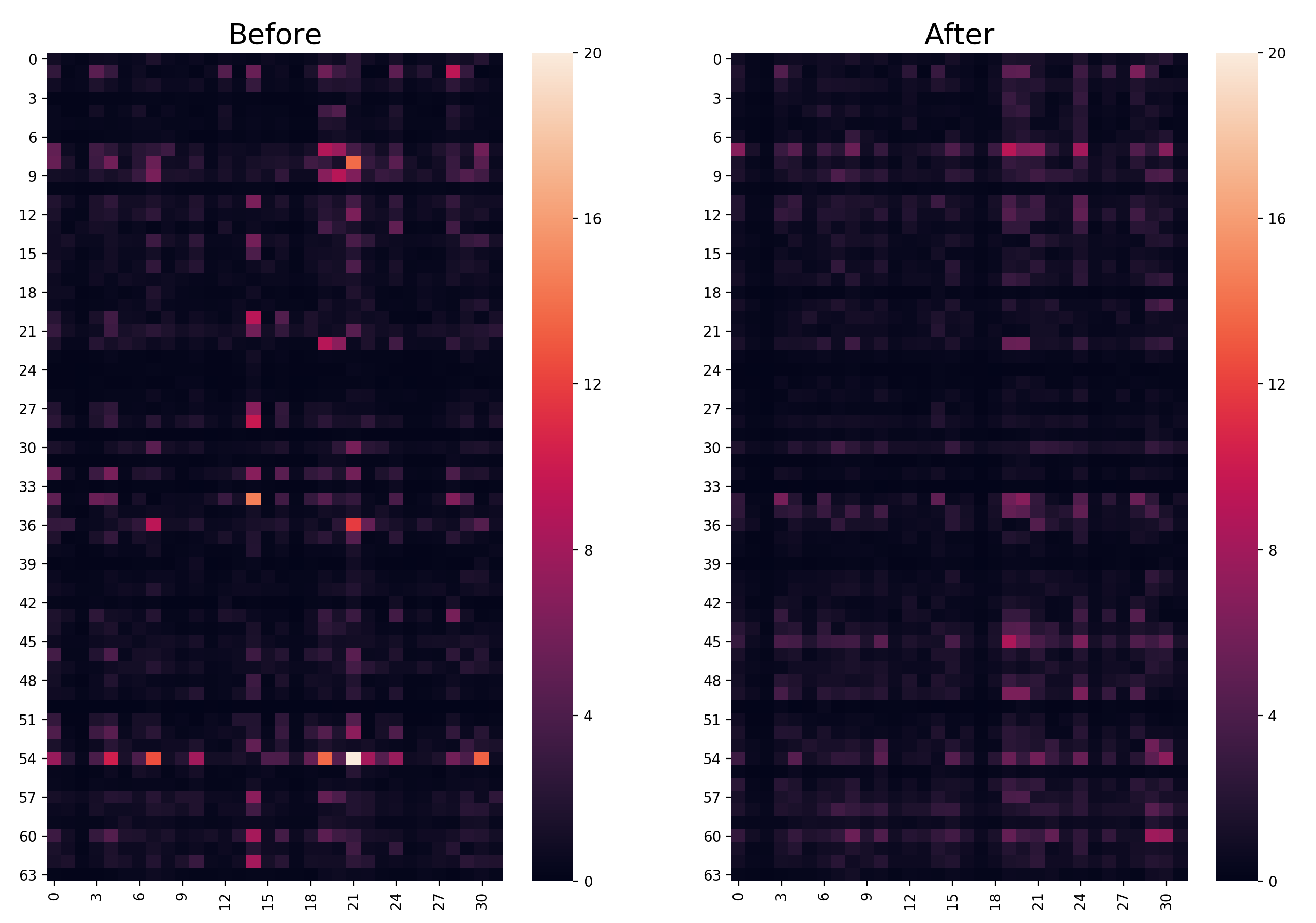}
    }
    \subfigure[En-Fr $\mathbf{W}_{\mathrm{E}_2}^{\mathrm{FFN}}$]{
        \includegraphics[width=0.3\columnwidth]{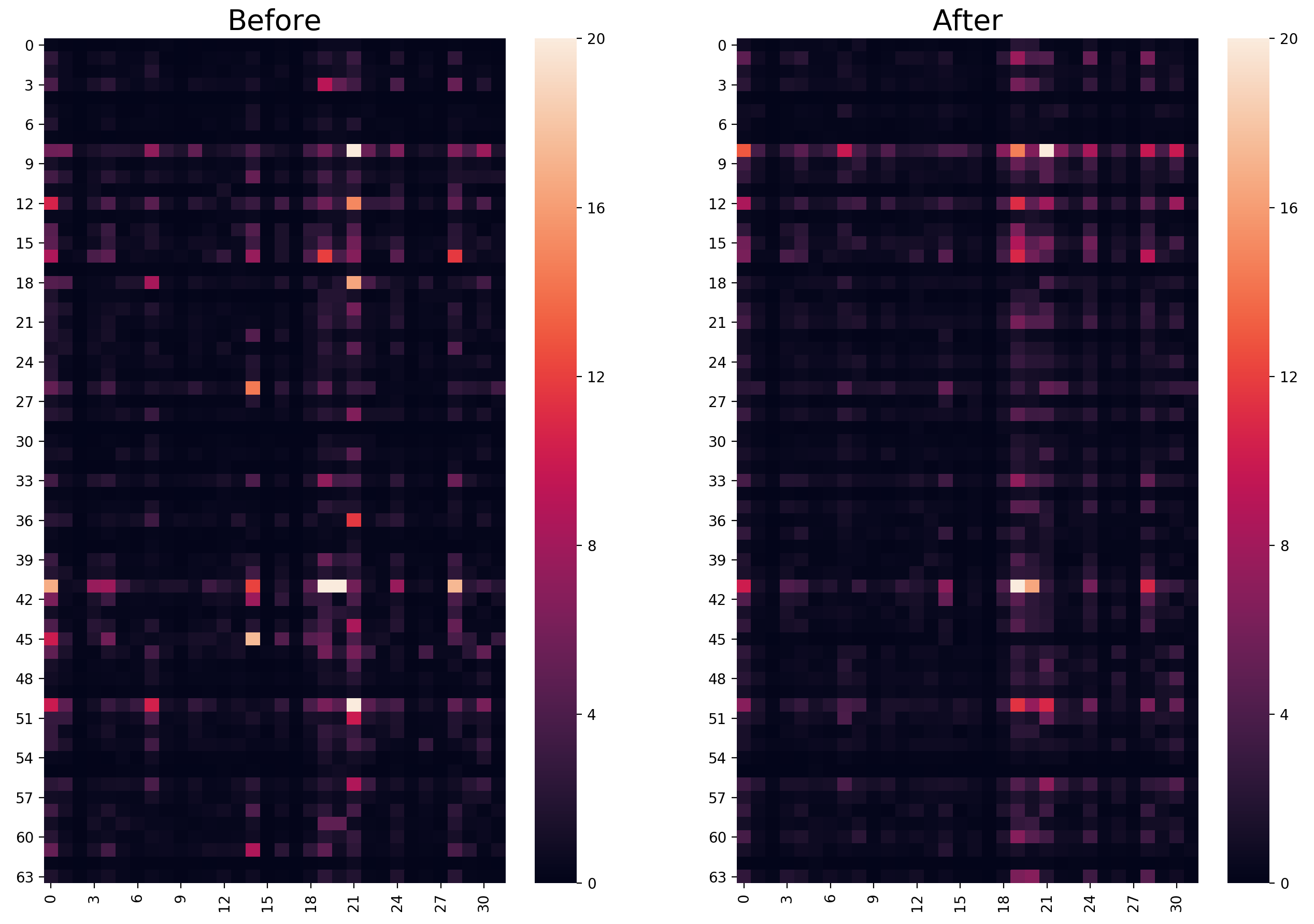}
    }
    \subfigure[En-De $\mathbf{W}_{\mathrm{D}_0}^{\mathrm{CA}}$]{
        \includegraphics[width=0.3\columnwidth]{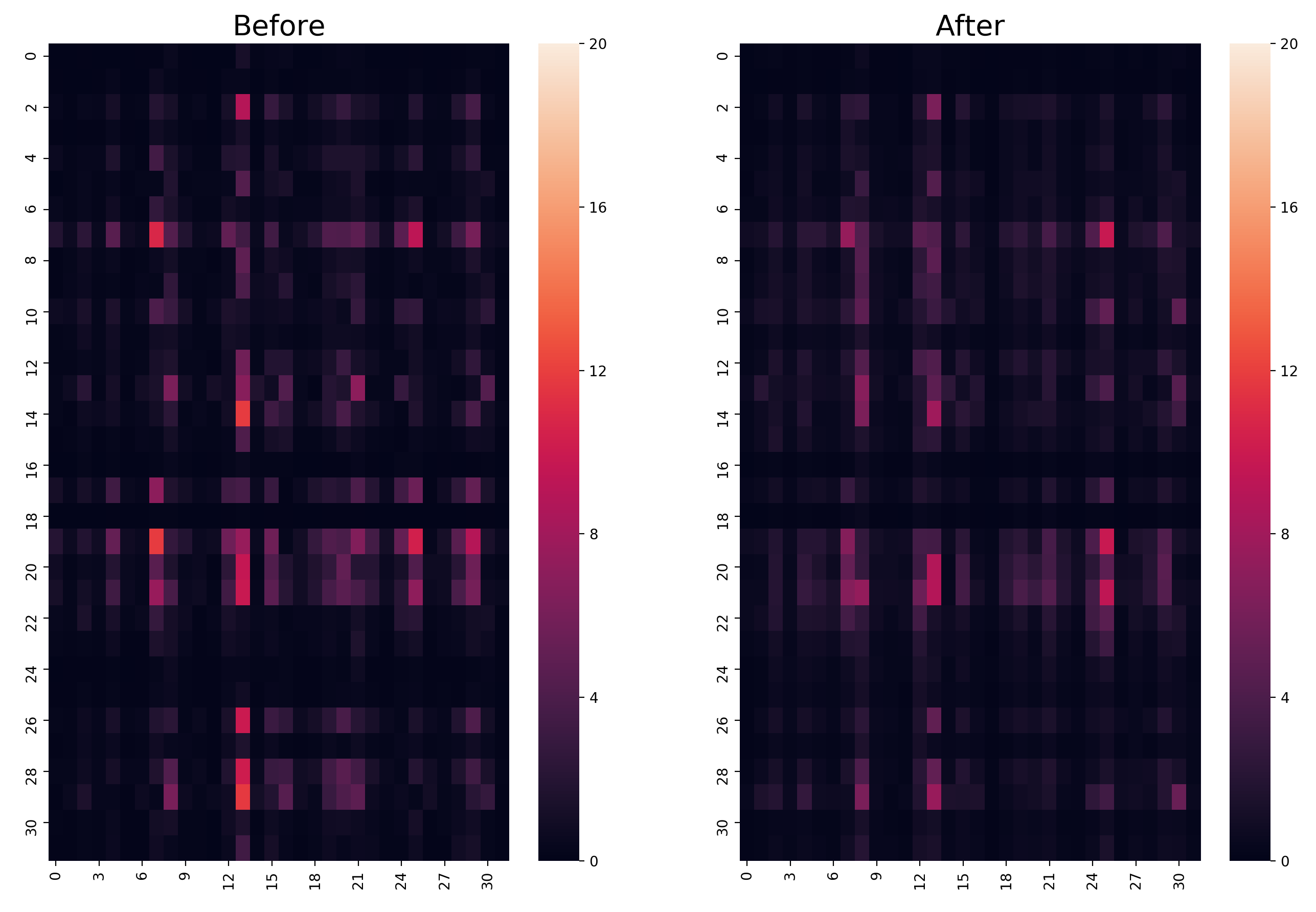}
    }
    \subfigure[Zh-En $\mathbf{W}_{\mathrm{D}_1}^{\mathrm{CA}}$]{
        \includegraphics[width=0.3\columnwidth]{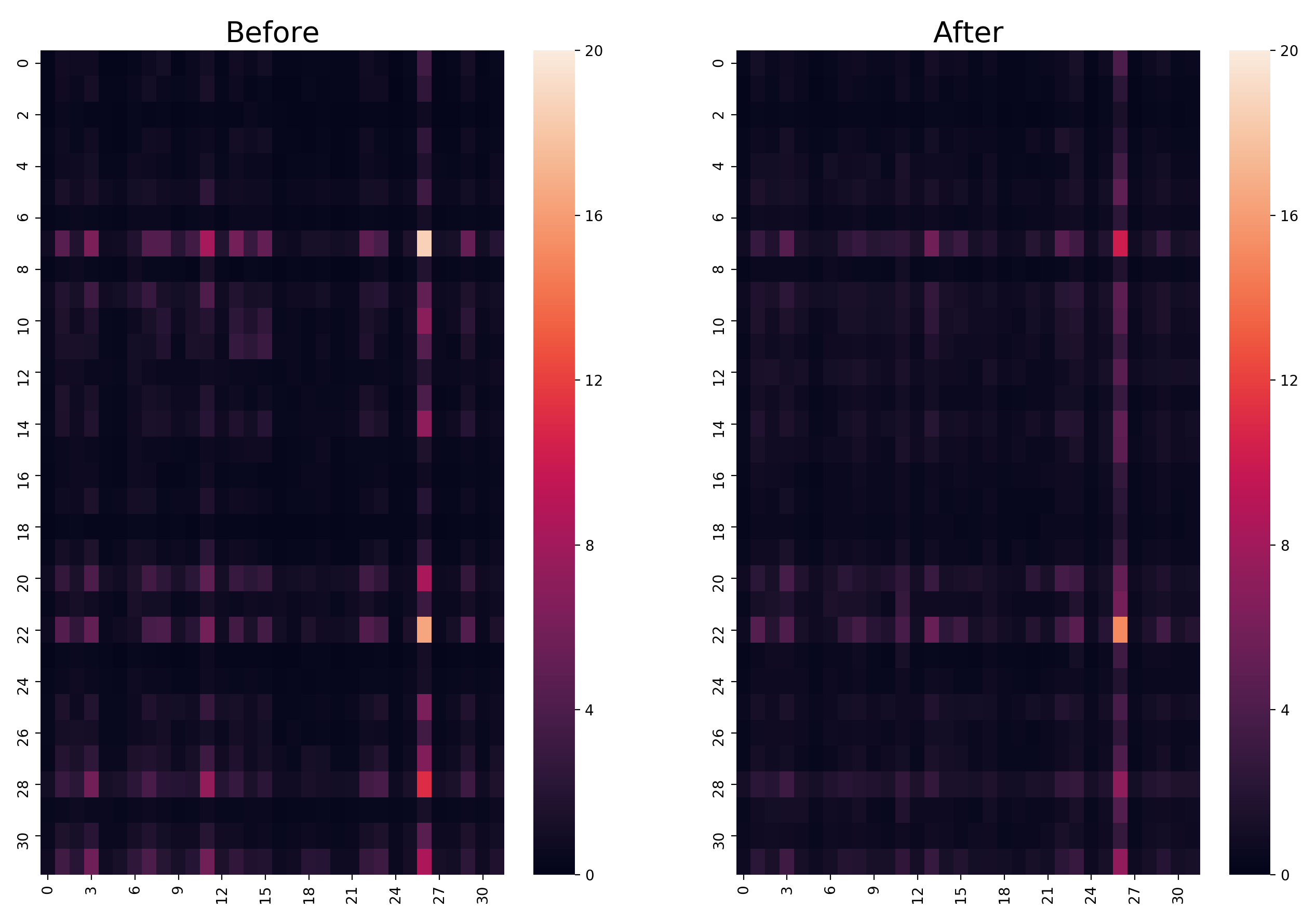}
    }
    \subfigure[En-Fr $\mathbf{W}_{\mathrm{D}_0}^{\mathrm{SA}}$]{
        \includegraphics[width=0.3\columnwidth]{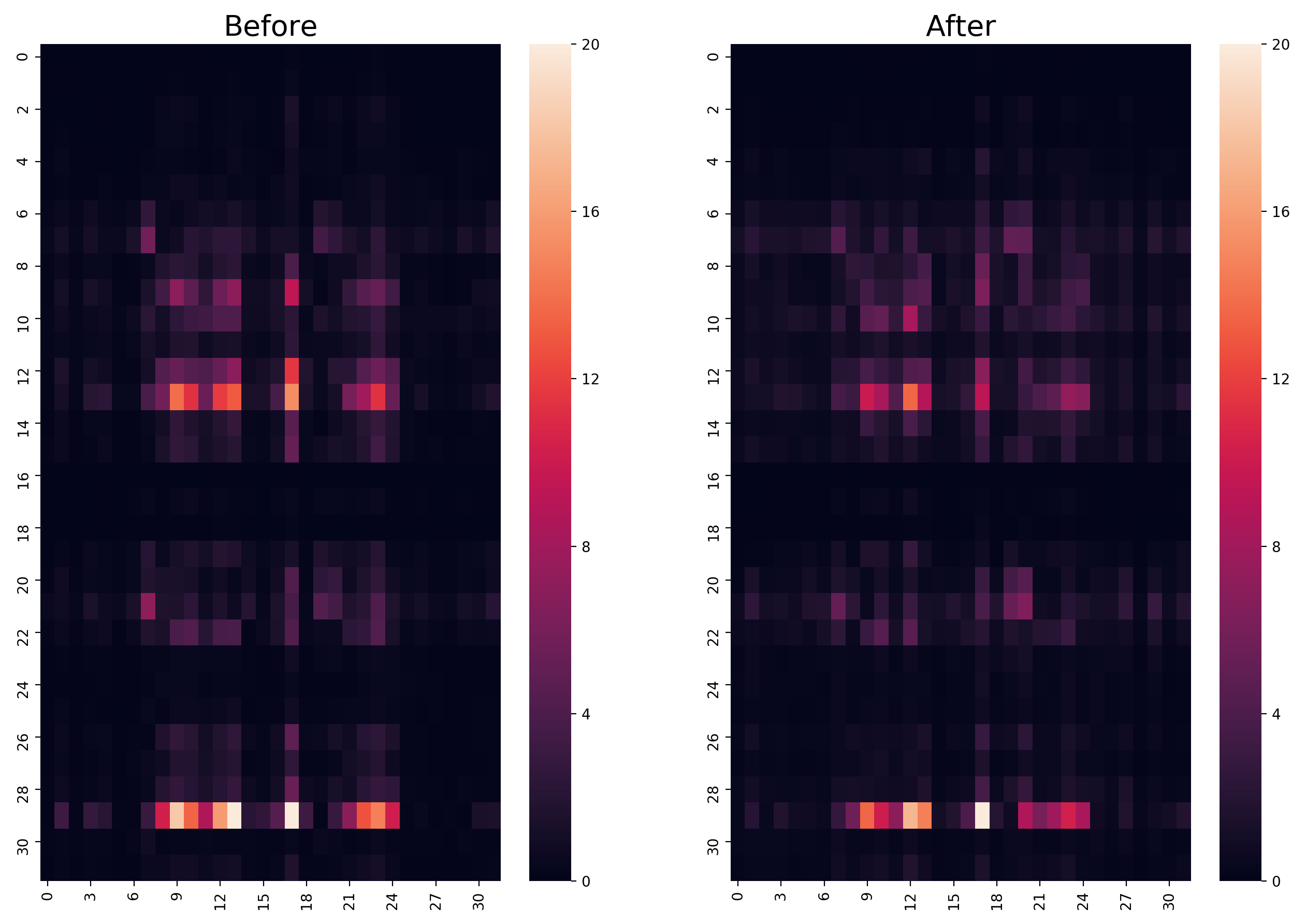}
    }
    \subfigure[En-De $\mathbf{W}_{\mathrm{D}_2}^{\mathrm{FFN}}$]{
        \includegraphics[width=0.3\columnwidth]{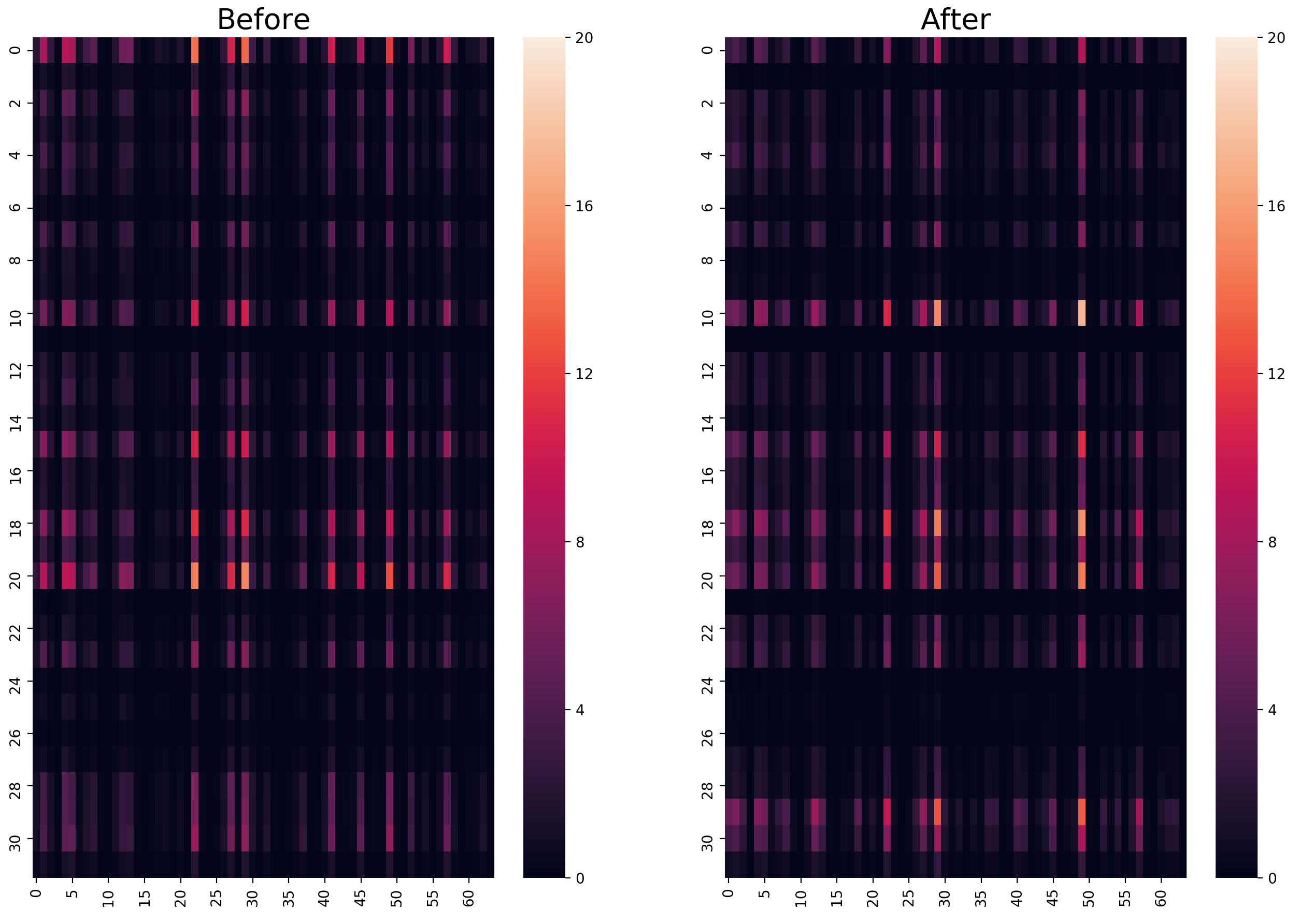}
    }

    \caption{Illustration of parameter importance before and after continual training. Each subfigure represents a parameter matrix of the model, where each square in it represents a parameter. The lighter the square in color, the more important the parameter. }
    \label{fig:heatmap}
\end{figure}

\subsection{Parameter Erasure} 
To prove that different parameters are indeed different in importance and verify the effectiveness of the proposed criterion for evaluating the importance of parameters, we conduct parameter erasure experiments to see the change of BLEU scores. 
For each parameter matrix $\mathbf{W}$ of the model, we rank all the parameters in it according to the proposed criterion and get the ranked list. Then we erase the model parameters (i.e., masking them to zero) increasingly according to the ranked list in ascending or descending order. If the evaluation results are correct, erasing the parameters in the descending order will hurt the translation quality more than erasing in the ascending order. 

\subsection{Experiments}
The training data and experimental systems are just the same as in section~\ref{exp}. 
\subsubsection{Parameter Erasure for the General-Domain Model}
Figure~\ref{fig:erasure1} shows part of the results of the parameter erasure experiments for all the three language pairs on the general-domain test sets. In most cases, not surprisingly, both of the two curves decrease monotonically and the curve of erasing the parameters in ascending order according to the importance is higher than the other one.
Based on this, we can conclude that some parameters are indeed more important than others for the whole model and have a larger impact on translation quality. 
In some cases, however, we also found some abnormal results: the two curves intersect or the curve is not monotonous. Although the independence of each parameter has been assumed during the derivation of the proposed method, it isn't always this case in practice, which will lead to these abnormal results. Overall, the proposed criterion can identify those important parameters in most cases, so that we can make use of it to analyze the behavior of the parameters during the continual training process.  

\subsubsection{Parameter Importance Visualization} 
The distribution of parameter importance can be seen as an important feature, which can imply the inner change of the model.
Therefore we try to visualize the importance distribution before and after continual training. To achieve this, we retrain models with the tiny parameter setting, considering the convenience for presentation. The model is first trained with the general-domain data and we get the importance evaluation matrices for all the modules, which are computed using the general-domain data. Then the model is continually trained with the in-domain data and we also get the importance evaluation matrices for the in-domain, which, in contrast, are computed with the in-domain data. These two matrices are then visualized with heatmaps. 

Figure~\ref{fig:heatmap} shows the results. 
Firstly, the more important parameters, which are lighter in the figures, lie in certain rows or columns of the target parameter matrix. Considering that the parameters in the same row or column are connected to the same neurons in the former or latter layer, we argue that some neurons are more important for the model, which is consistent with the conclusion of~\newcite{BauBSDDG19}. 
Secondly, the parameter distribution before and after continual training is very similar; most of the lighter squares in the left pictures are still lighter than other squares in the right pictures.
This observation result indicates that the parameters which are important for the general-domain translation still have larger impacts on the in-domain translation after the continual learning process in most cases. 
Lastly, there are still lots of less important parameters after continual training, which have limited influence on the in-domain translation. 

  \begin{figure}[t]
    \centering
    \subfigure[Zh-En]{
        \includegraphics[width=0.3\columnwidth]{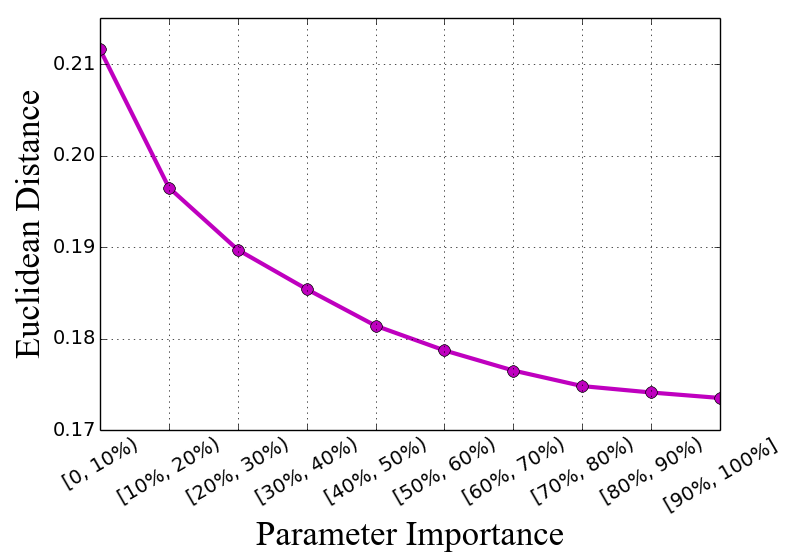}
    }
    \subfigure[En-Fr]{
        \includegraphics[width=0.3\columnwidth]{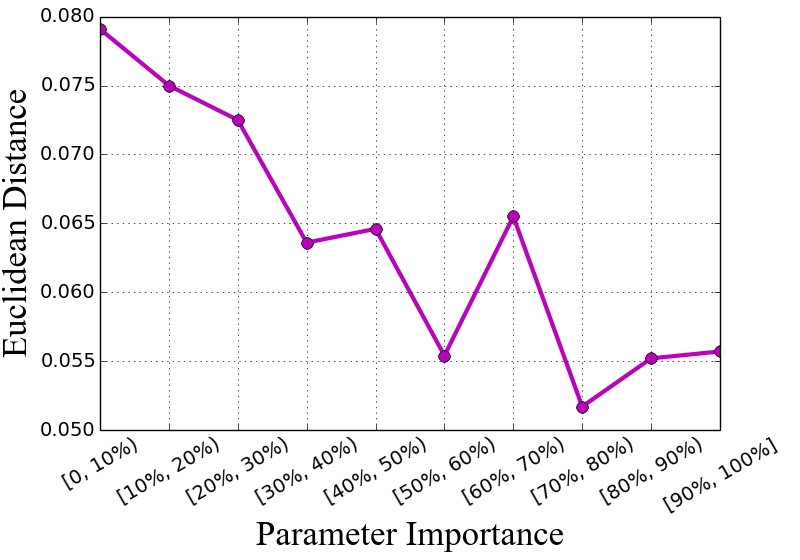}
    }
    \subfigure[En-De]{
        \includegraphics[width=0.3\columnwidth]{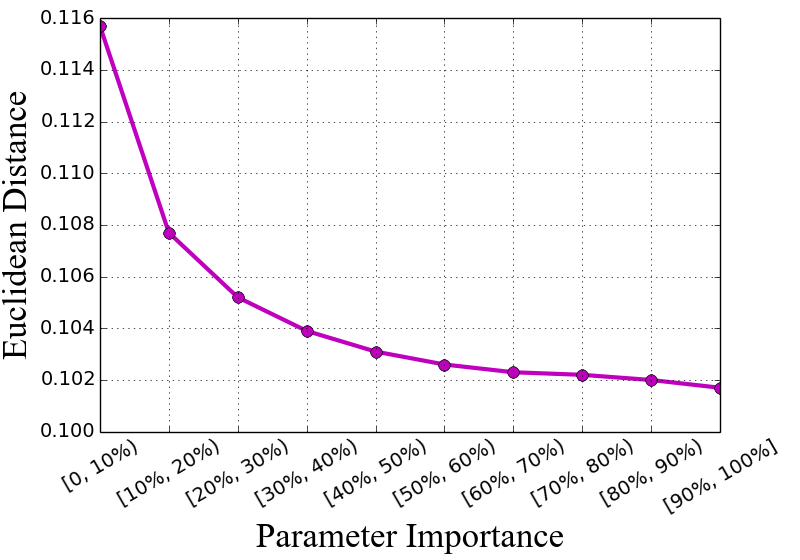}
    }
    \caption{The value change of the model parameters. The x-axis stands for the importance interval of the model before fine-tuning, e.g., $\left[20\%,30\%\right)$ means the importance of parameters is between top 20\% to 30\%. The y-axis represents the Euclidean distance between the parameters of the model before and after continual training.}
    \label{fig:distance}
\end{figure}


\subsubsection{Parameter Variation across Domains}
From the results above, we find that the important parameters for the general-domain still have large impacts on the in-domain translation after continual training. To figure out the variation of parameters with different importance, we compute the average Euclidean distance for all the parameters in the model before and after the continual learning process:
\begin{equation}
    \mathrm{distance}=\frac{1}{N} \sum_{i=1}^{N}\sqrt{(\mathbf{W}_{i}^\mathrm{G}-\mathbf{W}_{i}^\mathrm{I})^2}, 
\end{equation}
where $N$ denotes the number of different modules in the model; $\mathbf{W}_{i}$ denotes the parameter matrix for the $i$-th module; $\mathrm{G}$ and $\mathrm{I}$ denote the general-domain and in-domain, respectively. Then all the parameters in each module are ranked and divided into ten groups according to their importance. The result of the average Euclidean distance of parameters in each importance interval is shown in Figure~\ref{fig:distance}. We find that the top parameters change more greatly than the less important parameters. Considering their impacts on the translation, we conclude that it is because of the excessive change of the important parameters that causes the catastrophic forgetting phenomenon.



\subsection{Summary and Inspiration}
In this section, we propose a method for evaluating the parameter importance. Then through the parameter erasure experiments, we find that some parameters are more important and have a greater influence on the output. Next based on the importance distribution visualization results, we find that some parameters are important for both of the domains.
Finally, the average Euclidean distance of model parameters before and after continual learning is calculated and we find that the important parameters change more greatly, which causes the catastrophic forgetting problem. Inspired by our findings, we can freeze part of those important parameters during continual training to avoid catastrophic forgetting. Besides, we can retrain those unimportant parameters to further improve the in-domain translation.

\section{Related Work}
\textbf{Analyzing Work} Recently, much work has been concerned with analyzing and evaluating the NMT model from different perspectives.
\newcite{ShiKY16} investigates how NMT models output target strings of appropriate lengths. \newcite{ding-etal-2017-visualizing} analyzes the contribution of each contextual word to arbitrary hidden states. \newcite{QiSFPN18} analyzes when the pre-trained word embeddings can help in NMT tasks. \newcite{VoitaTMST19} analyzes the importance of different attention heads. \newcite{BauBSDDG19} investigates the importance and function of different neurons in NMT. 
\newcite{WuebkerSD18} finds that a large proportion of model parameters can be frozen during adaptation with
minimal reduction in translation quality by encouraging structured sparsity.
\newcite{WangS20} links the exposure bias problem~\cite{RanzatoCAZ15,ShaoCF18,ZhangFMYL19} to the phenomenon of NMT tends to generate hallucinations under domain shift.
\newcite{abs-2010-04380} finds that the NMT tends to generate more high-frequency tokens and less low-frequency tokens than reference.
Compared with them, this work mainly focuses on investigating the functions of the different modules and parameters in the NMT model during continual training. 
In this sense, the work of~\newcite{ThompsonKAMDMMG18} is most related to ours, which tries to understand the effectiveness of continued training for improving the in-domain performance. Compared with their work, our work also explores the cause of the catastrophic forgetting problem in general-domain. Besides, our work also analyzes the performance of NMT at the neuron (parameter) level. 

\textbf{Continual Training} Continual training, which is also referred to as fine-tuning, is widely used in NMT for the domain adaptation task. 
\newcite{luong2015stanford} fine tunes the general-domain model with the in-domain data. 
\newcite{chu2017empirical} fine tunes the model with the mix of the general-domain data and over-sampled in-domain data. \newcite{KhayrallahTDK18} and \newcite{ThompsonGKDK19} add regularization terms to let the model parameters stay close to their original values. \newcite{dakwale2017fine} minimizes the cross-entropy between the output distribution of the general-domain model and the fine-tuned model. \newcite{GuFL19} adds a discriminator to help preserve the domain-shared features and fine tunes the whole model on the mixed training data. 
\newcite{JiangLWZ20} proposes to obtain the word representations by mixing their embedding in individual domains based on the domain proportions.
\newcite{abs-2010-04003} presents a theoretical analysis of catastrophic forgetting in the Neural Tangent Kernel regime.
Compared with them, our work pays attention to exploring the inner change of the model during continual training as well as the cause of the catastrophic forgetting phenomenon.

\section{Conclusion}
In this work, we focus on the catastrophic forgetting phenomenon of NMT and aim to find the inner reasons for this. Under the background of domain adaptation, we propose two analyzing methods from the perspectives of modules and parameters (neurons) and conduct experiments across different language pairs and domains. We find that some modules tend to maintain the general-domain knowledge while some modules tend to adapt to the in-domain; we also find that some parameters are more important for both the general-domain and in-domain translation and the change of them brings about the performance decline in general-domain. Based on our findings, we have proposed several ideas that may help improve the vanilla continual training method. We will prove the effectiveness of these ideas in future work.



\section*{Acknowledgements}
We thank all the anonymous reviewers for their insightful and valuable comments. This work was supported by National Key R\&D Program of China (NO. 2017YFE0192900).

\bibliographystyle{coling}
\bibliography{coling2020}

\end{document}